\documentclass[conference,11pt]{IEEEtran}
\IEEEoverridecommandlockouts
\usepackage[a4paper,left=1.5cm,right=2.1cm,top=1.9cm,bottom=4.4cm]{geometry}
\usepackage[table]{xcolor}
\usepackage{cite}
\usepackage{amsmath,amssymb,amsfonts}
\usepackage{array}
\usepackage{url}
\usepackage{algorithmic}
\usepackage{graphicx}
\usepackage{textcomp}
\usepackage{xcolor}
\usepackage{hyperref}
\hypersetup{
    colorlinks=true,
    urlcolor=blue,
    citecolor=black
}

\def\BibTeX{{\rm B\kern-.05em{\sc i\kern-.025em b}\kern-.08em
		T\kern-.1667em\lower.7ex\hbox{E}\kern-.125emX}}
\begin{document}
	\title{Real-Time Threat Detection from Surveillance
Cameras using Machine Learning}
	\author{
\IEEEauthorblockN{Gajendra Mandal, Dr.~J.~P.~Patra, Priyansh Mahant}
\IEEEauthorblockA{
\textit{Department of Computer Science \& Engineering} \\
\textit{Chhattisgarh Swami Vivekanand Technical University}, Bhilai, India \\
gajendra.300012822016@csvtu.ac.in, jppatra.cse@csvtu.ac.in, priyanshmahant200@gmail.com
}
}
	\maketitle
   
	\begin{abstract}
Ensuring public safety in densely populated urban environments remains a critical challenge, necessitating the deployment of intelligent and automated video surveillance systems. Traditional surveillance approaches rely heavily on manual monitoring, which is inefficient and susceptible to human fatigue, delayed response, and observational errors. To overcome these limitations, this work presents a real-time object detection-based surveillance framework.

The proposed system focuses on detecting guns, knives, and region-specific blunt objects commonly involved in violent activities in Indian surveillance scenarios. A key contribution of this work is the use of a custom-created dataset collected using a mobile camera, consisting of 336 labeled images of blunt objects such as iron rods, wooden sticks, and plastic rods. This dataset is combined with a publicly available dataset of 7,623 images of guns and knives, forming a consolidated dataset of 7,959 images across three classes: gun, knife, and blunt\_object.

The combined dataset is used to train a YOLOv8-based object detection model for real-time performance. Experimental evaluation shows that increasing the training duration significantly improves recall and average precision for the blunt\_object class without signs of overfitting. Overall, the proposed framework achieves an effective balance between accuracy and efficiency, making it suitable for deployment in real-world surveillance environments such as campuses, public spaces, and transportation areas.
\end{abstract}

	\renewcommand{\IEEEkeywordsname}{Keywords}

\section{Introduction}

Urban infrastructural development has necessitated an increase in the use of
video surveillance cameras for public safety. The traditional monitoring systems
require human intervention in monitoring, making them vulnerable to factors
like human error, fatigue, etc. Smart surveillance systems, using deep
learning and computer vision techniques, can be considered more efficient
alternatives~\cite{b1,b2}.

Among the various methods of object detection, YOLO offers good real-time
performance for detection of weapons in videos~\cite{b1}. Although
surveillance systems employing object detection technique have been widely
used to detect firearms and edged weapons, they rely upon non-Indian
databases~\cite{b3,b4}. However, blunt weapons like iron rods, wooden sticks,
and plastic rods are commonly found in violent events in India, and they are
highly unrepresented in the current databases and detection systems~\cite{b4}.

In order to bridge this gap, we present a surveillance method using YOLOv8
with enhanced capabilities to detect guns, knives, and region-specific blunt
objects. A custom image database of 336 images featuring different types of
blunt objects, captured in Indian scenes, was collected and combined with a
publicly available gun and knife database to obtain a consolidated dataset
comprising 7,959 images belonging to three classes~\cite{mandal_combined_dataset}.
Results reveal that extended training helps improve the recall and AP score for
the blunt objects while avoiding any kind of overfitting. Moreover, the
method proves to be real-time feasible.

Our major contributions include:
\begin{itemize}
  \item Identifying the gap present in current weapon datasets with respect to Indian
        surveillance environment.
  \item A region-specific custom dataset containing images of blunt weapons~\cite{mandal_blunt_dataset}.
  \item Implementation and evaluation of real-time three-class detector using YOLOv8.
  \item Extended training helps avoid overfitting~\cite{mandal_combined_dataset}.
\end{itemize}

\section{Related Work}

 Previous weapon detection systems have concentrated mainly on detecting
guns and edged weapons through deep learning algorithms. Singh and Mohan
have employed deep spatio-temporal learning for road surveillance
\cite{b3}, and Ingle and Kim \cite{b4} have expanded
object detection for smart cities. Nevertheless, both these papers use
foreign data sources that do not include Indian region-specific blunt
objects. Qi et al. \cite{qi2021} released a large-scale gun detection
dataset comprising more than 51{,}000 images and an edge-cloud gun
alerting framework; however, their algorithm was only limited to detecting
guns. On the other hand, Thakur et al. \cite{thakur2024} and Deshpande et
al. \cite{deshpande2023} have used YOLOv8 to demonstrate weapon
detection in public areas, proving the applicability of real-time weapon
detection in public places but omitting blunt weapons and behavioral
analysis.

In terms of violence detection, researchers have stressed the need for
temporal modeling, with the CNN--LSTM combination performing
significantly well on all datasets \cite{violencereview2024}. Based on the
above research, the proposed work combines a temporal frame buffer and a
two-stage vision-language model (VLM) inference mechanism for
context-aware threat assessment \cite{zhang2024vlm}, providing
structured security reports instead of mere bounding box detection.

Table~\ref{tab:comparison} summarizes the key differences between existing
systems and the proposed framework.
 
\begin{table}[htbp]
\caption{Comparison Between Existing Systems and Proposed Framework}
\label{tab:comparison}
\centering
\renewcommand{\arraystretch}{1.3}
\resizebox{\columnwidth}{!}{%
\begin{tabular}{|l|c|c|c|c|c|c|c|}
\hline
\textbf{Work} &
\textbf{\shortstack{Gun}} &
\textbf{\shortstack{Knife}} &
\textbf{\shortstack{Blunt\\Obj.}} &
\textbf{\shortstack{Indian\\Data}} &
\textbf{\shortstack{Behavior\\Analysis}} &
\textbf{\shortstack{Alert}} &
\textbf{\shortstack{Threat\\Level}} \\
\hline
Qi et al.~\cite{qi2021}               & \checkmark & $\times$   & $\times$   & $\times$   & $\times$   & \checkmark & $\times$   \\
Thakur et al.~\cite{thakur2024}       & \checkmark & \checkmark & $\times$   & $\times$   & $\times$   & $\times$   & $\times$   \\
Deshpande et al.~\cite{deshpande2023} & \checkmark & $\times$   & $\times$   & $\times$   & $\times$   & $\times$   & $\times$   \\
Ingle \& Kim~\cite{b4}         & \checkmark & \checkmark & $\times$   & $\times$   & $\times$   & $\times$   & $\times$   \\
\hline
\textbf{Proposed}                     & \checkmark & \checkmark & \checkmark & \checkmark & \checkmark & \checkmark & \checkmark \\
\hline
\end{tabular}%
}
\end{table}

\section{Proposed work}
	This section describes the proposed work for real-time intelligent surveillance, focusing on the overall system methodology and the data collection strategy adopted to address limitations in existing datasets. The proposed system combines deep learning–based object detection with vision–language–based behavioral analysis to achieve accurate, context-aware, and interpretable threat detection in live video streams.
    
    \subsection{Methodology}\label{AA2} A proposed system methodology is depicted in Fig. 1 below. The suggested system applies an event-driven methodology that includes real-time object detection, temporal context modeling, and vision-language model (VLM)-based behavioral reasoning to detect potential threats efficiently.

Real-time object detection in the proposed methodology starts with live video capturing from a surveillance camera working with the frequency of 30 FPS. The frames acquired during video capturing are then processed in the preprocessing stage that involves image resizing, normalization, and frame rate adjustment. A two-stream approach is used to detect objects; the high-frequency frames are used to detect objects, while the frames captured at low frequency are used to analyze behaviors. 

YOLOv8 based model is applied to perform object detection to identify any potentially dangerous objects like guns, knives, or blunt objects. The output from the model consists of bounding boxes, object classes, and confidence levels. Only the detections with confidence higher than some threshold value will be recognized as potential threats. Otherwise, the system returns back to the surveillance mode~\cite{b1}. 

Upon detecting a threat, the decision gate will activate the behavior analysis process. First, the cooldown mechanism will apply to avoid any redundant behavior analyses of the already recognized suspicious objects. Besides, a temporal frame buffer will store selected frames (approx. 5s) before, during, and after object detection. This temporal buffer can help in analyzing object behavior in more detail, as well as detecting its activities accurately. 

The next step in analyzing the suspicious object is the two-stage analysis of behavioral patterns with the Vision-Language Model (VLM). At the first stage, rapid screening will be performed with several key frames from the temporal buffer. This stage works as an optimization technique to filter out non-suspicious behavior. If suspicious behavior is confirmed at this stage, the second one is triggered. The second stage includes a detailed behavior analysis and a generation of human readable and understandable security reports about actions, threat severity, etc.

\begin{figure}[htbp]
    \centering
    \includegraphics[width=0.48\textwidth, height=12cm, keepaspectratio=false]{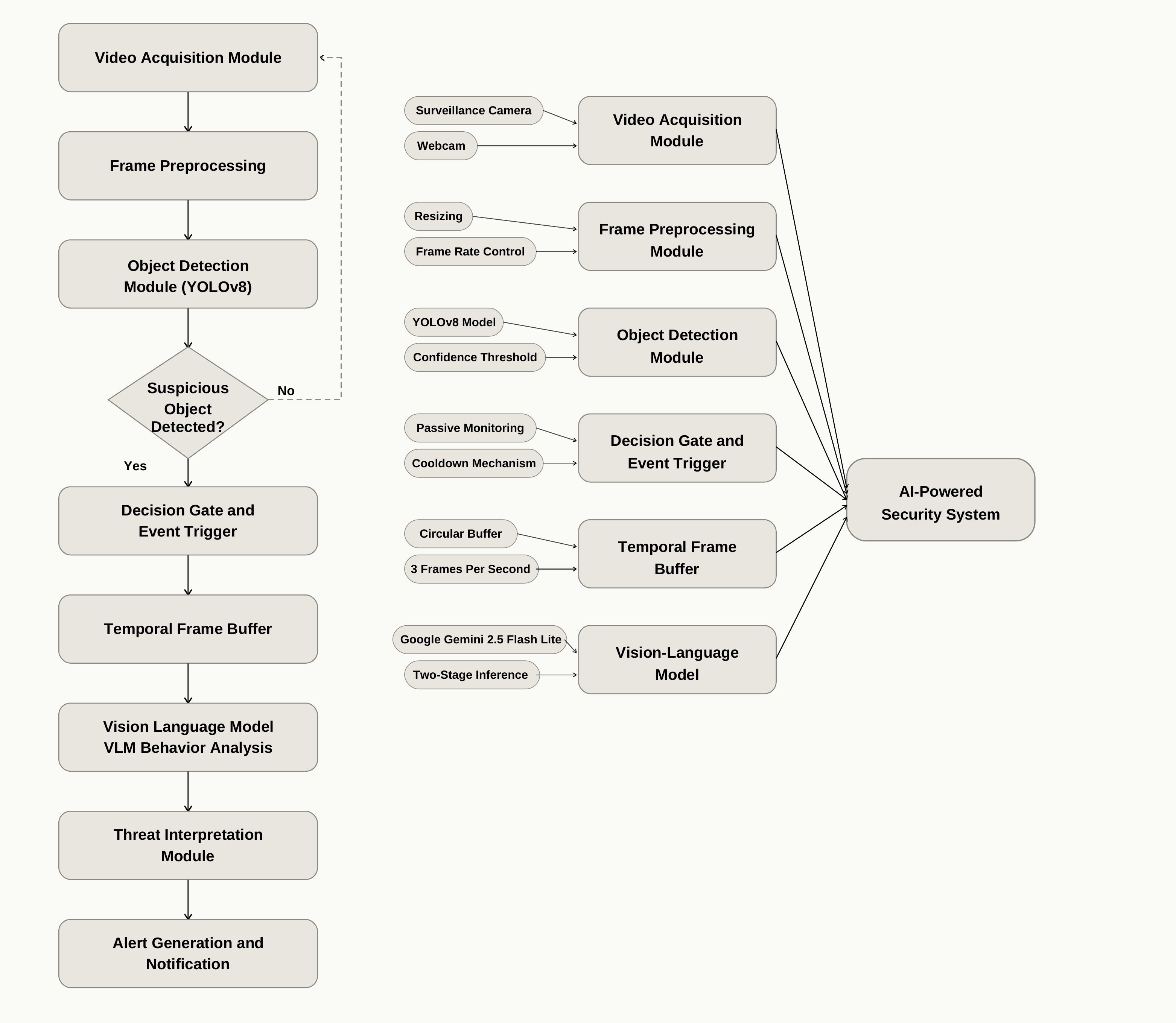}
    \caption{Workflow of the proposed methodology}
    \label{fig:workflow}
\end{figure}

Finally, the detected objects and their behaviors are analyzed further by threat interpretation logic. On this stage, all gathered information is processed by the threat classification algorithm that sorts the detected behavior threats in categories CRITICAL, HIGH, MEDIUM, and LOW. Depending on the classification results, automatic alerts with appropriate security reports will be distributed via email, desktop notifications, and system log files.
\ref{fig:workflow}. 
    
  \subsection{Data Collection}\label{AA}
One major limitation found in the existing datasets related to weapon detection is that the Indian public environment is poorly represented. Publicly available weapon detection datasets comprise non-Indian public scenes and mainly focus on firearms and knives. In an effort to overcome this issue, this study presents a customized dataset consisting of region-specific blunt weapons, which is publicly available for future research purposes~\cite{mandal_blunt_dataset}.

\subsubsection{Custom Blunt Object Dataset}

This dataset has been curated by the authors using a smartphone-based approach in order to add realism. The videos have been recorded using a \textit{Realme GT Master Edition} smartphone and cover different indoor and outdoor scenes with varying degrees of background illumination, background clutter, camera angle, and object orientation. A total of 15 scenes have been recorded using the mentioned device and with a frame size of $1080 \times 1920$, recorded at 60 FPS.

From the recorded videos, 336 different frames have been selected manually in order to avoid duplication. These frames contain different types of region-specific blunt objects and are annotated using bounding boxes, which are compliant with the YOLO annotation requirements. These annotated frames belong to three types of blunt objects, namely, \textit{iron\_rod}, \textit{wooden\_rod}, and \textit{plastic\_rod}. All annotated frames belong to the Indian context and have been publicly available through~\cite{mandal_blunt_dataset}.

\subsubsection{Public Dataset for Guns and Knives}

In order to obtain robust results when detecting firearms and knives, this paper utilizes a publicly available dataset from the Roboflow Universe. The dataset used is called the \textit{Gun and Knife Detection} dataset~\cite{roboflow_gun_knife}, which is provided by Mahad Ahmed.

\subsubsection{Final Dataset Composition}
The final training dataset is a combination of the custom blunt object dataset~\cite{mandal_blunt_dataset} and the public gun and knife dataset~\cite{roboflow_gun_knife}. The consolidated dataset consists of three object classes: \textit{gun}, \textit{knife}, and \textit{blunt\_object}. In total, the dataset contains 7,959 annotated images, with all images resized to a uniform resolution of $640 \times 640$ pixels to ensure consistency during training and inference.
    
\section{Results and Discussion}

This section presents the experimental evaluation of the proposed YOLOv8-based object detection model and discusses its performance in terms of training duration, class-wise detection accuracy, and alignment with the defined objectives. The analysis focuses exclusively on object detection performance to justify the final model selection for real-time intelligent surveillance deployment.

\subsection{Experimental Setup and Evaluation Protocol}

The object detection model was trained on a consolidated dataset consisting of 7,959 annotated images across three object classes: \textit{gun}, \textit{knife}, and \textit{blunt\_object}. All images were resized to a uniform resolution of $640 \times 640$ pixels prior to training.

Training and evaluation were performed using GPU acceleration on an NVIDIA GeForce GTX 1650 (4~GB VRAM). To study the impact of training duration, the model was evaluated at two checkpoints: after 50 epochs and after 100 epochs. Evaluation was conducted on a validation set containing 1,588 images with 1,755 object instances.

Performance was measured using standard object detection metrics, including Precision (P), Recall (R), mean Average Precision at IoU threshold 0.5 (mAP@0.5), and mean Average Precision averaged over IoU thresholds from 0.5 to 0.95 (mAP@0.5:0.95), following the COCO evaluation protocol. Precision–Recall curves and confusion matrices were obtained directly from the YOLOv8 validation pipeline.

\subsection{Epoch-wise Performance Comparison}

Table~\ref{tab:epoch_comparison} summarizes the quantitative performance comparison between 50 and 100 training epochs, including the observed changes in key metrics.

\begin{table}[htbp]
\caption{Performance Comparison Between 50 and 100 Epochs}
\label{tab:epoch_comparison}
\centering
\resizebox{\columnwidth}{!}{%
\begin{tabular}{lccc}
\hline
\textbf{Metric} & \textbf{50 Epochs} & \textbf{100 Epochs} & \textbf{Change} \\
\hline
Precision (All)      & 0.862 & 0.857 & --0.005 \\
Recall (All)         & 0.676 & \textbf{0.757} & \textbf{+0.081} \\
mAP@0.5 (All)        & 0.777 & \textbf{0.819} & \textbf{+0.042} \\
mAP@0.5:0.95 (All)   & 0.556 & \textbf{0.570} & \textbf{+0.014} \\
Gun mAP@0.5          & 0.905 & \textbf{0.919} & +0.014 \\
Knife mAP@0.5        & 0.767 & \textbf{0.785} & +0.018 \\
Blunt Object mAP@0.5 & 0.658 & \textbf{0.754} & \textbf{+0.096} \\
\hline
\end{tabular}%
}
\end{table}

The results indicate that extending training to 100 epochs improves recall and mean Average Precision across all classes, while precision remains largely stable, demonstrating improved detection sensitivity without a significant increase in false positives.

\subsection{Discussion of Results}

The most significant performance enhancement is seen in recall, where there is an 8.1\% increase in performance after 100 epochs. Recall is especially important for any kind of surveillance application because failure to detect threatening objects could result in life-threatening consequences.

The global mAP@0.5 also increases by 4.2\%, proving that localization and object classification performance improve with more training. The relatively low increase in mAP@0.5:0.95 shows that there is a slow increase in localization accuracy.

The precision score is also slightly lower, suggesting that the model is not suffering from overfitting.

\subsection{Class-wise Performance Analysis}

\begin{figure}[htbp]
\centering
\includegraphics[width=0.45\textwidth]{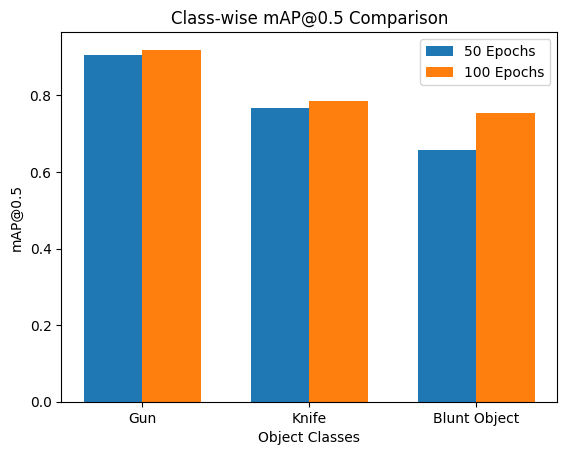}
\caption{Class-wise mAP@0.5 comparison between 50 and 100 epochs for gun, knife, and blunt\_object classes.}
\label{fig:classwise_map}
\end{figure}

Figure~\ref{fig:classwise_map} shows the class-wise mAP@0.5 comparison between 50 and 100 epochs. The gun class achieves the highest performance, reaching an mAP@0.5 of 0.919 at 100 epochs, owing to consistent visual structure and sufficient training samples.

Knife detection improves moderately with extended training, while the blunt\_object class exhibits the largest relative gain (9.6\%). This improvement highlights the effectiveness of longer training for learning region-specific objects with high intra-class variability.

\subsection{Epoch-wise Convergence Analysis}

\begin{figure}[htbp]
\centering
\includegraphics[width=0.45\textwidth]{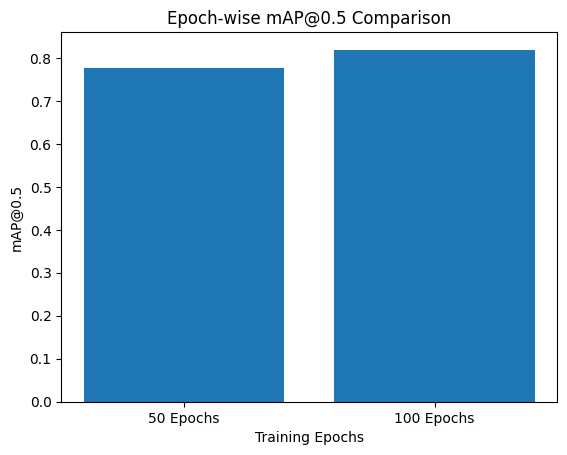}
\caption{Epoch-wise comparison of overall mAP@0.5 showing improved convergence at 100 epochs.}
\label{fig:epoch_map}
\end{figure}

Figure~\ref{fig:epoch_map} presents the epoch-wise comparison of overall mAP@0.5. The increase from 0.777 at 50 epochs to 0.819 at 100 epochs demonstrates continued learning and improved convergence with extended training.

\subsection{Precision-Recall Curves and Confusion Matrix}

\begin{figure}[htbp]
\centering
\includegraphics[width=0.50\textwidth]{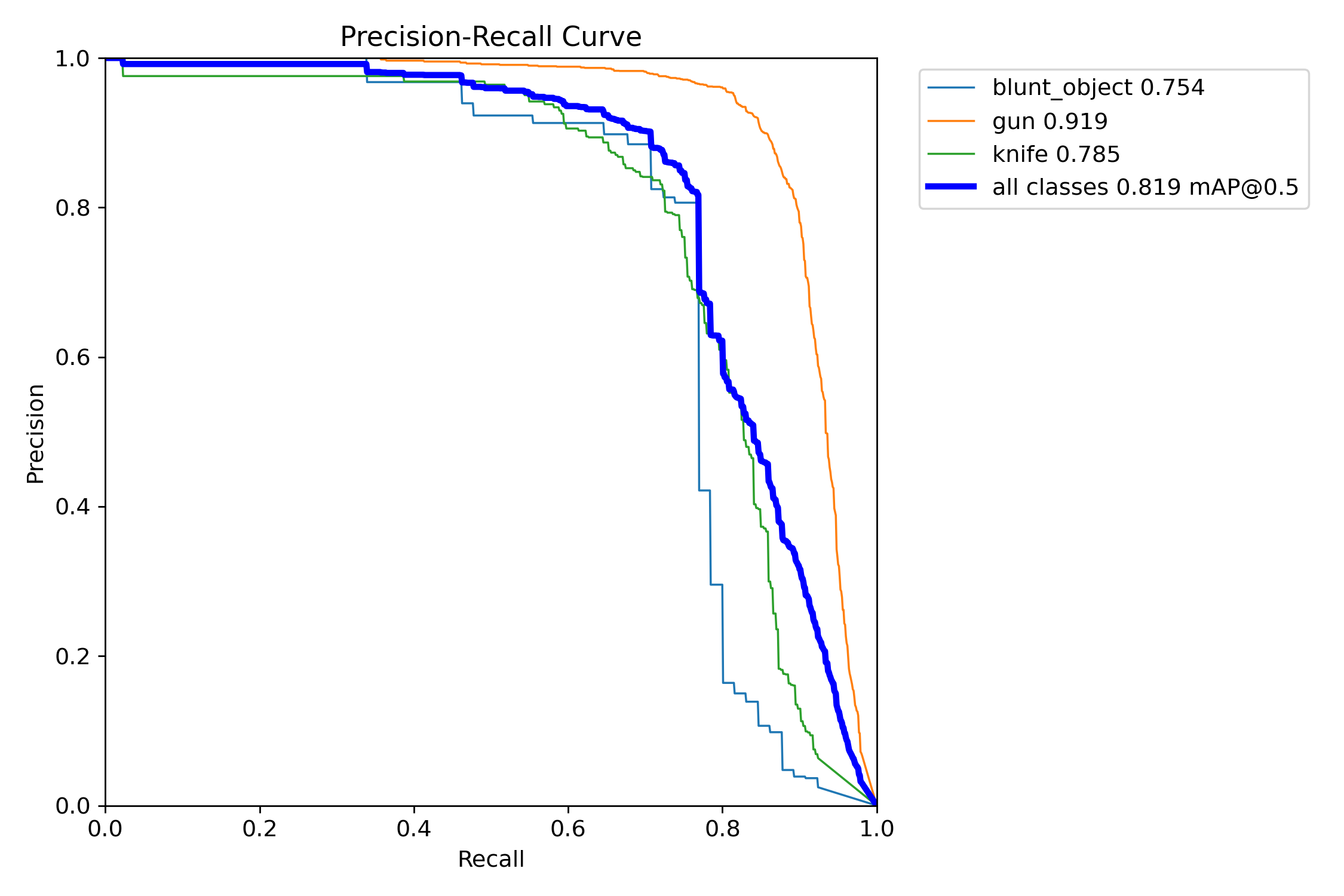}
\caption{Precision-Recall curves for object detection classes obtained from YOLOv8 validation of the final model trained for 100 epochs.}
\label{fig:pr_curve}
\end{figure}

Figure~\ref{fig:pr_curve} presents the Precision--Recall curves obtained from the validation of the \textit{final selected model trained for 100 epochs}. The curves illustrate the trade-off between precision and recall for different object classes and demonstrate stable precision across a wide recall range. In particular, the gun and knife classes maintain high precision even at higher recall levels, indicating reliable detection performance for safety-critical objects.

\begin{figure}[htbp]
\centering
\includegraphics[width=0.54\textwidth]{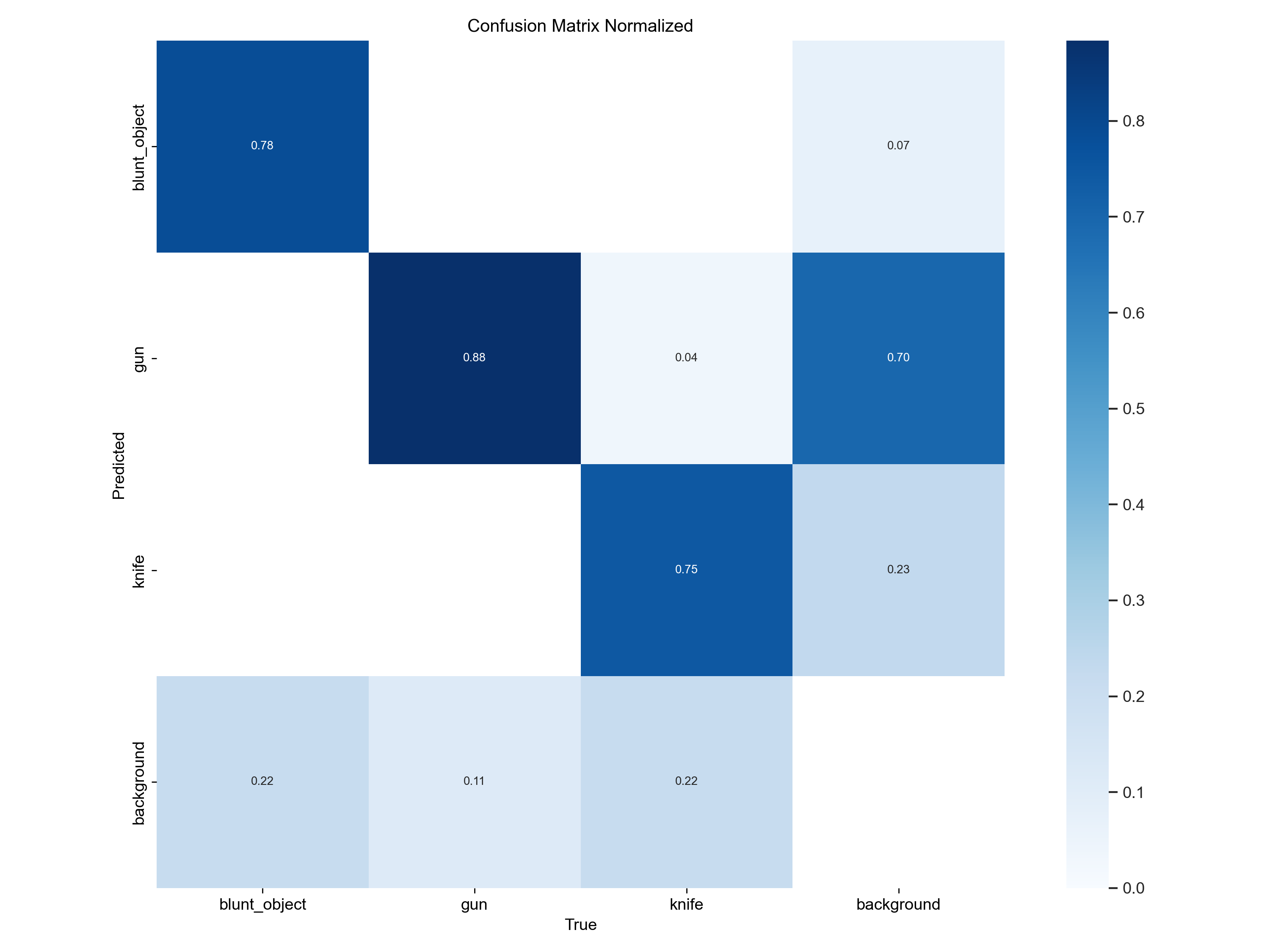}
\caption{Confusion matrix illustrating class-wise prediction performance on the validation dataset for the final selected model (100 epochs).}
\label{fig:conf_matrix}
\end{figure}

Figure~\ref{fig:conf_matrix} shows the confusion matrix corresponding to the same final model. The strong diagonal dominance confirms correct class predictions for the majority of validation samples. The remaining misclassifications primarily involve visually ambiguous instances of the blunt\_object class, which is expected due to its high intra-class variability and diverse appearance.

\subsection{Results and Objectives Alignment}

Table~\ref{tab:objective_alignment} presents the alignment between the defined objectives and the achieved experimental results.

\begin{table}[htbp]
\caption{Alignment Between Objectives and Experimental Results}
\label{tab:objective_alignment}
\centering
\renewcommand{\arraystretch}{1.25}
\resizebox{\columnwidth}{!}{%
\begin{tabular}{>{\raggedright\arraybackslash}p{3.2cm}
                >{\raggedright\arraybackslash}p{3.5cm}
                >{\raggedright\arraybackslash}p{3.5cm}}
\hline
\textbf{Objective} & \textbf{Method Used} & \textbf{Outcome} \\
\hline
Detect firearms in surveillance
& YOLOv8 object detection
& Gun mAP@0.5 = 0.919 \\

Detect knives reliably
& YOLOv8 object detection
& Knife mAP@0.5 = 0.785 \\

Detect region-specific blunt objects
& Custom dataset + YOLOv8
& Blunt object mAP@0.5 improved by 9.6\% \\

Improve recall for safety-critical use
& Extended training (100 epochs)
& Recall improved by 8.1\% \\

Maintain real-time feasibility
& Lightweight YOLOv8 model
& Stable precision and fast inference \\
\hline
\end{tabular}%
}
\end{table}

\subsection{Model Selection Justification}

Based on the experimental results, the model trained for 100 epochs is selected as the final model for deployment. This decision is supported by improved recall, higher mean Average Precision, significant gains for the blunt\_object class, and stable precision without signs of overfitting.

\subsection{Qualitative Evaluation and System Deployment}

\begin{figure}[htbp]
\centering
\includegraphics[width=0.48\textwidth]{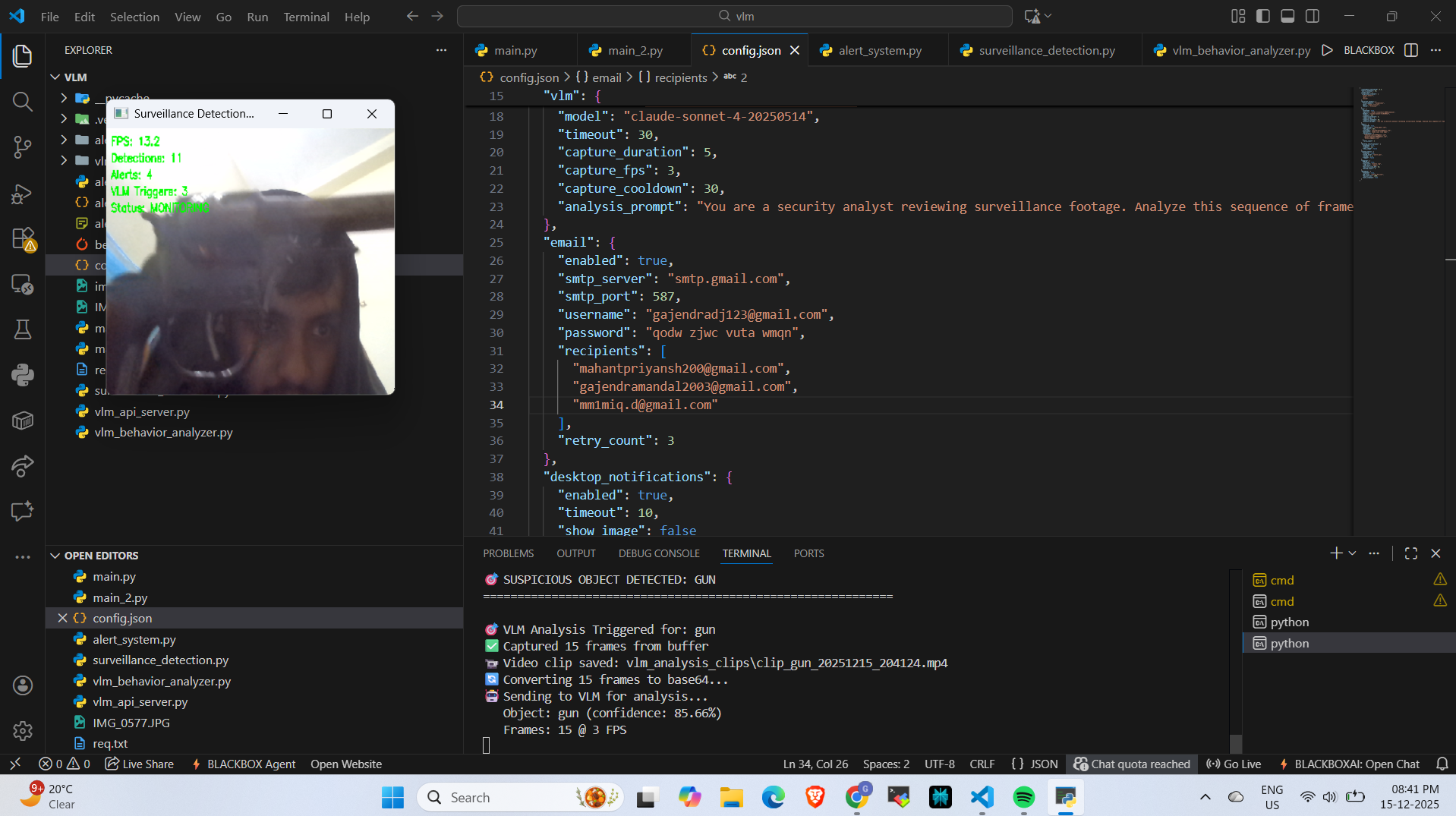}
\caption{Real-time testing of the proposed surveillance pipeline showing live camera input, object detection output, system status indicators, and runtime logs during inference.}
\label{fig:pipeline_testing}
\end{figure}

To further validate the practical feasibility of the proposed surveillance framework, a qualitative evaluation was conducted during real-time system testing. Figure~\ref{fig:pipeline_testing} presents a snapshot of the system during live execution, demonstrating the end-to-end operation of the surveillance pipeline under realistic conditions.

As shown in Fig.~\ref{fig:pipeline_testing}, the system processes live camera input and performs object detection in real time, overlaying detection results directly on the video stream. Information like FPS, confidence value, and system status, among others, which show that the system is under constant monitoring, are displayed in the visual interface. At the same time, the terminal shows that all internal processing, including triggers from detection, frame caching, and completion of inference, is being performed correctly by the system.

It is evident that the above result indicates that the proposed system works as expected in a well-integrated system, performing video capture, object detection, decision-making, and alerts. It also confirms that the system does not function only as a theoretical one that works off-line, making it more practically relevant to intelligent surveillance systems.

\subsection{Summary of Findings}

Through the results obtained from the experiment, it is clear that the performance in terms of object detection has been greatly improved by increased training sessions. In addition to this, the proposed model provides good accuracy on firearms and knives, with great improvements seen in specific regions for blunt objects.
	\section*{Acknowledgment}
	This work was conducted as a minor project at Chhattisgarh Swami Vivekanand Technical University (CSVTU), Bhilai. The authors thank CSVTU for institutional support and research guidance. The authors also acknowledge the use of AI-based language models for minor assistance in editing, proofreading, and formatting of the manuscript. The research work, methodology, and results are entirely the authors' own.
    
	\section*{}

	\vspace{12pt}
	
\end{document}